\pdfoutput=1

\documentclass[11pt]{article}

\usepackage{EACL2023}

\usepackage{times}
\usepackage{latexsym}

\usepackage[T1]{fontenc}

\usepackage[utf8]{inputenc}

\usepackage{microtype}

\usepackage{inconsolata}

\usepackage{microtype}
\usepackage{mathrsfs}
\usepackage{amsmath}
\usepackage{amssymb}
\usepackage{graphicx}
\usepackage{algorithm}
\usepackage{algpseudocode}
\usepackage{multirow}
\usepackage{booktabs}
\usepackage{url}
\usepackage{CJKutf8}
\usepackage{adjustbox}
\usepackage{enumitem}
\usepackage{lipsum}
\usepackage{xcolor,colortbl}
\usepackage{caption}
\usepackage{subcaption}
\usepackage{soul}

\title{Which One Are You Referring To? \\ Multimodal Object Identification in Situated Dialogue}

\author{Holy Lovenia\thanks{\hspace{1.5mm}Equal contribution.}\hspace{1.5mm}, Samuel Cahyawijaya$^*$, Pascale Fung \\
  Center for Artificial Intelligence Research (CAiRE), \\
  The Hong Kong University of Science and Technology \\
  \texttt{\{hlovenia, scahyawijaya\}@connect.ust.hk} \\}

\begin{document}
\maketitle
\begin{abstract}
The demand for multimodal dialogue systems has been rising in various domains, emphasizing the importance of interpreting multimodal inputs from conversational and situational contexts.
One main challenge in multimodal dialogue understanding is multimodal object identification, which constitutes the ability to identify objects relevant to a multimodal user-system conversation.
We explore three methods to tackle this problem and evaluate them on the largest situated dialogue dataset, SIMMC 2.1. Our best method, scene-dialogue alignment, improves the performance by $\sim$20\% F1-score compared to the SIMMC 2.1 baselines. We provide analysis and discussion regarding the limitation of our methods and the potential directions for future works. Our code is publicly available at \url{https://github.com/holylovenia/multimodal-object-identification}.


\end{abstract}

\section{Introduction}

Recent advancements in multimodal dialogue systems have gained more traction in various domains such as retail, travel, fashion, interior design, and many others. A real-world application of multimodal dialogue systems is situated dialogue, where a dialogue agent shares a co-observed vision or physical space with the user, and is responsible for handling user requests based on the situational context, which are often about the objects in their surroundings. This makes multimodal object identification from a dialogue (i.e., identifying objects that fit a dialogue context) an indispensable skill in multimodal dialogue understanding, built on cross-modal understanding to comprehend the relations between linguistic expressions and visual cues.

Various methods have been proposed to perform multimodal object identification through different paradigms~\cite{yu2016modeling, hu2016natural, ilinykh-etal-2019-tell, aishwarya2021mdetr, kuo2022beyond}.
These efforts have established remarkable progress in solving this problem. However, aside from an observed gap between the performance of the existing works and human-level performance in multimodal object identification, prior works also rely on a presumption that the information given by the textual context will only lead to specific (i.e., unambiguous) objects, which does not conform to real-world multimodal conversations where ambiguity exists.
%

\begin{figure}
    \centering
    \resizebox{\linewidth}{!}{
    \includegraphics{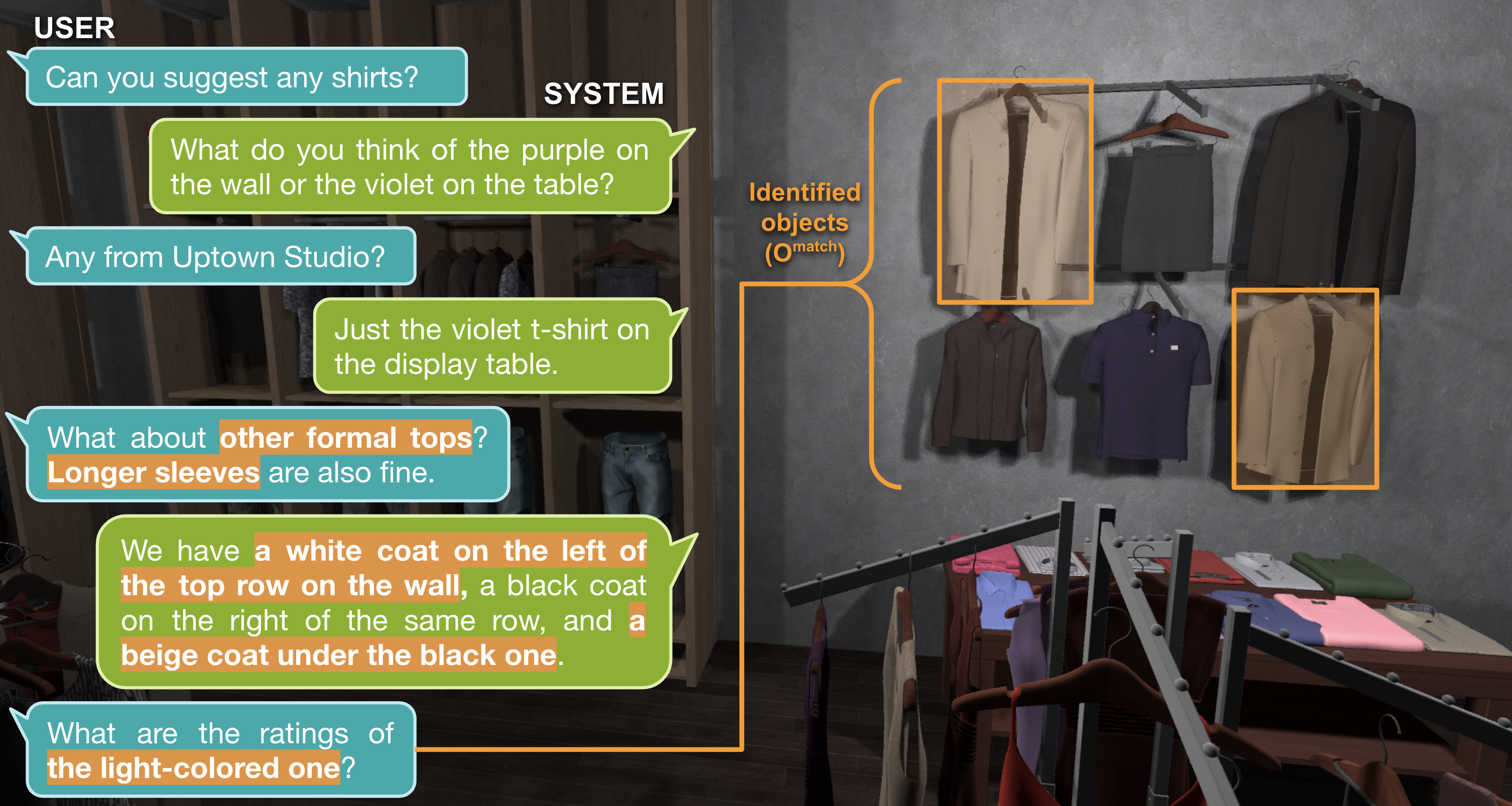}
    }
    \caption{Multimodal object identification is the fundamental step required to enable multimodal dialogue systems to understand the object referred to by the user. Image is adapted from~\cite{kottur2021simmc}.
    }
    \label{fig:overview}
\end{figure}

Therefore, in this work,
we explore three different solutions to enable multimodal object identification in the situated dialogue system, i.e., dialogue-contextualized object detection, object-dialogue alignment, and scene-dialogue alignment, without adopting the unambiguity assumption. Dialogue-contextualized object detection utilizes the spatial and object understanding capability of a pre-trained object detection model, to generate semantic representation containing both visual cues and the spatial understanding of the object. Object-dialogue alignment incorporates the image-text alignment capability of CLIP~\cite{radford2021clip}, which has been pre-trained on large image-text corpora to perform multimodal object identification from the given dialogue context. Scene-object alignment combines the spatial and object understanding capability of a pre-trained object detection model and a pre-trained textual understanding model to produce better semantic vision-language alignment.

Our contributions are three-fold:
\begin{itemize}
    \item We introduce three different methods for handling multimodal object identification in situated dialogue, i.e., 
    dialogue-contextualized object detection, object-dialogue alignment, and scene-dialogue alignment;
    \item We show the dialogue-contextualized object detection method fails to outperform even the heuristic baselines despite having an acceptable performance on the object detection task;
    \item We show the effectiveness of the other two methods which significantly outperform the SIMMC 2.1 baselines by $\sim$5\% F1-score for object-dialogue alignment and $\sim$20\% F1-score for scene-dialogue alignment;
\end{itemize}

\section{Related Work}

\paragraph{Multimodal Dialogue System}

Multiple studies have attempted to enable the skills required for multimodal dialogue system, e.g., understanding visual~\cite{antol2015vqa, das2017visual, kottur-etal-2019-clevr} or visual-temporal~\cite{alamri2019audio} content to answer user's questions, grounding conversations to images~\cite{mostafazadeh-etal-2017-image, shuster-etal-2020-image}, interpreting multimodal inputs and responding with multimodal output to assist users with their goal~\cite{saha2018towards} or as a means to converse~\cite{sun-etal-2022-multimodal}, and perceiving the shared environment to grasp situational context to enable proper navigation, adaptation, and communication~\cite{lukin-etal-2018-scoutbot, brawer2018situated, kottur2021simmc}.

At the core of these efforts, the ability to understand language and vision, as well as integrate both representations to align the linguistic expressions in the dialogue with the relevant visual concepts or perceived objects, is the key to multimodal dialogue understanding~\cite{landragin2006visual, loaiciga-etal-2021-reference, loaiciga-etal-2021-annotating, kottur2018visual, utescher2021did, sundar-heck-2022-multimodal, dai-etal-2021-multimodal}.

\begin{figure*}[t!]
    \centering
    \resizebox{0.9\linewidth}{!}{
        \includegraphics{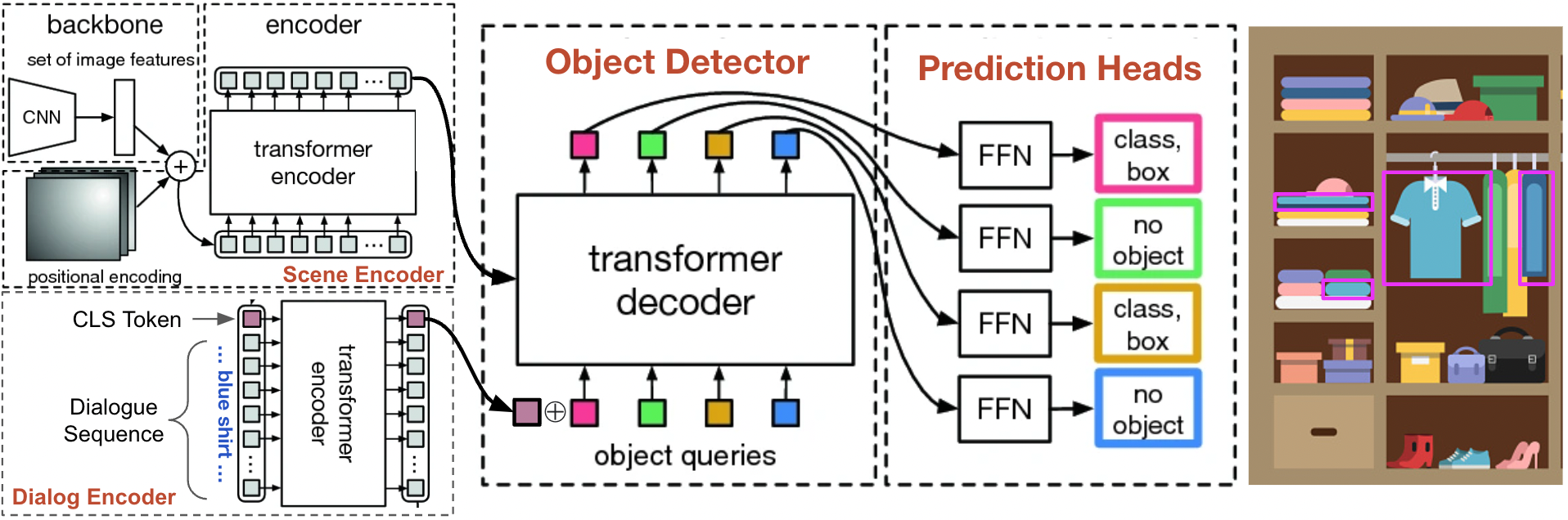}
    }
    \caption{The architecture of SitCoM-DETR. SitCoM-DETR consists of a scene encoder and a dialogue encoder to extract multimodal content, respectively. The dialogue representation is used to guide the object detector module to judiciously filter out unrelated scene objects.}
    \label{fig:sitcom-detr}
\end{figure*}

\paragraph{Multimodal Object Identification}

Identifying objects or visual concepts related to a linguistic expression is an incremental exploration in vision-language research. It starts with identifying simple objects in a sanitized environment~\cite{mitchell2010natural} based on image descriptions or captions. Then, multimodal object identification has been gradually increasing in complexity and realism by involving visual contexts with cluttered and diverse scenes~\cite{kazemzadeh-etal-2014-referitgame, gkatzia-etal-2015-virtual, yu2016modeling, mao2016generation, hu2016natural, ilinykh-etal-2019-tell, aishwarya2021mdetr, kuo2022beyond}.

While these works base their multimodal object identification on single-turn text contexts, another line of works explores the usage of multi-turn sequences as a textual context to enable identifying objects based on implicit constraints deduced through multi-round reasoning~\cite{seo2017visual, johnson2017clevr, liu2019clevr, moon-etal-2020-situated}. However, they focus on identifying only the specific (i.e., unambiguous) objects, in which only a certain object in the scene fits the corresponding linguistic context. This is quite dissimilar from real-world multimodal object identification, where multiple objects could fit a given textual context and induce ambiguity into the conversation~\cite{kottur2021simmc}. For this reason, existing works are not equipped with the ability to identify all objects that \textit{plausibly} fit those constraints although this skill is required to perform multimodal object identification in situated dialogue.


\paragraph{Multimodal and Cross-Modal Learning}

Past works have studied multimodal and cross-modal alignment, grounding, and generation to solve various vision-language tasks, e.g., image captioning~\cite{hossain2019comprehensive, sharma2018conceptual}, generating stories from image~\cite{min2021deep, lovenia-etal-2022-every}, as well as multimodal object identification~\cite{li2019zero, wang2022cris}. These attempts become more substantial and extensive after the rise of pre-trained vision-language models such as CLIP~\cite{radford2021clip}, ALIGN~\cite{jia2021scaling}, and FLAVA~\cite{singh2022flava}, which allows transfer knowledge obtained from the large-scale pre-training to downstream tasks.



\section{Methodology}
\label{sec:methods}

In this section, we describe the preliminaries of our work (\S\ref{sec:prelim}) and extensively elaborate on each of our approaches, i.e., dialogue-contextualized object detection (\S\ref{sec:contextualized-obj-det}), object-dialogue alignment (\S\ref{sec:obj-dialogue-align}), and scene-dialogue alignment (\S\ref{sec:scene-dialogue-align}).

\subsection{Preliminaries}
\label{sec:prelim}

The goal of multimodal object identification in situated dialogue is to identify objects from a given scene image that fulfill the user's request gathered from the user-system interactions. To identify the object(s) that could satisfy a user's request in a dialogue, it is crucial to match the objects and the implicit constraints interwoven in the dialogue, e.g., S: ``\textit{I do! Take a look at these. I have \ul{a brown coat towards the far end on the left wall}, \ul{another brown coat on the left side of the front floor rack}, and \ul{a black coat on the front of the same rack}.}'', U: ``\textit{Awesome! Tell me the cost and label on \ul{that one}.}''. Thus, it is essential for the system to understand the relation between the visual perception of the objects in the scenes and the natural language used to verbalize these constraints, which describe the target object(s) by visual attributes (e.g., color, object category or type, etc.), location (i.e., absolute or relative position), or the combination of both.

We define a dialogue between a user and a system as $D = \{u_1, s_1, u_2, s_2, \dots, u_n, s_n\}$, a scene consisting of images corresponding to multiple viewpoints of the scene as $\{I^{scene}_1, I^{scene}_2, \dots, I^{scene}_n\}$, and a set of objects in the scene as $O^{scene} = \{(b_1, c_1), (b_2, c_2), \dots, (b_n, c_n)\}$, where $u_i$ and $s_i$ respectively denote the user utterance and the system utterance, and $c_i$ and $b_i$ denote the bounding box and the class category of an object. Given a user dialogue turn $D^{user}_i = \{u_1, s_1, u_2, s_2, \dots, u_i\}$, $i \leq n$, and a scene image $I^{scene}_i$, the goal of the task is to select a subset of scene objects $O^{match} \subseteq O^{scene}$ that could satisfy the referred criteria in $D^{user}_i$.

\begin{figure*}[t!]
    \centering
    \includegraphics[width=0.9\linewidth, trim={0 0 0 0}, clip]{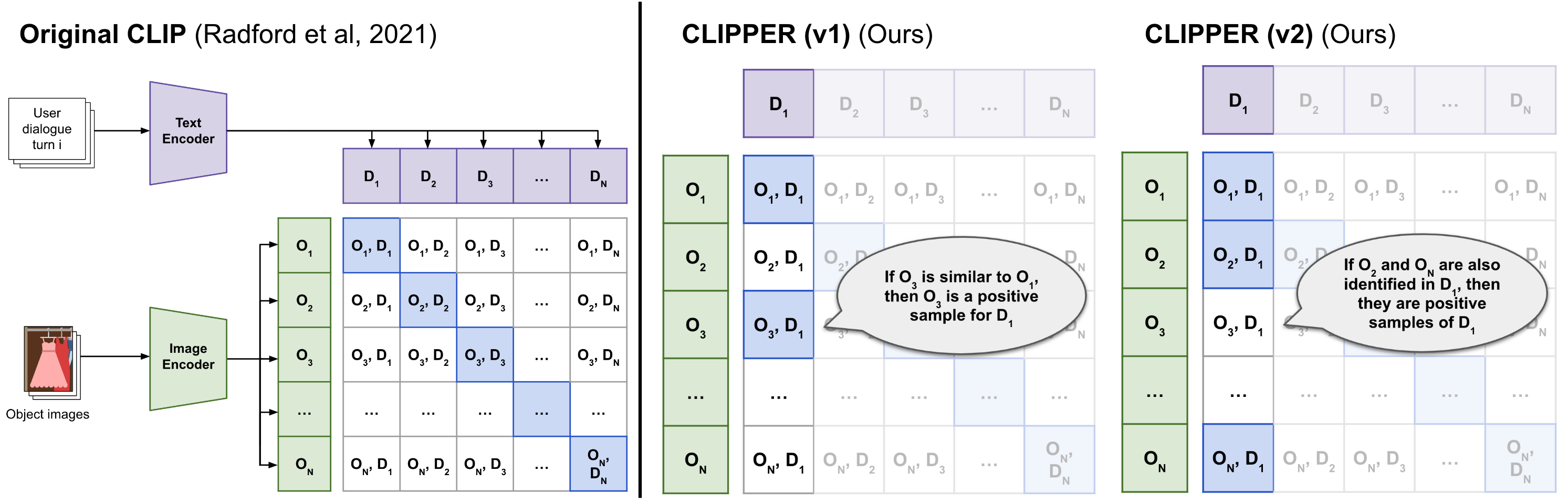}
    \caption{Learning objectives of the original CLIP~\cite{radford2021clip}, CLIPPER (v1), and CLIPPER (v2) for the object-dialogue alignment approach. The similarities of the positive pairs (blue) are maximized while the similarities of the negative pairs (white) are minimized.}
    \label{fig:contrastive-learning}
\end{figure*}

\subsection{Approach 1: Dialogue-Contextualized Object Detection}
\label{sec:contextualized-obj-det}

For dialogue-contextualized object detection, we frame the task of multimodal object identification as the contextualized object detection task. In object detection, given a scene image $I^{scene}$, we aim to detect all objects $O^{scene}$ in the scene by predicting their bounding box and class category. While in contextualized object detection, the aim is instead to select only a set of scene objects $O^{match}$ that satisfy a given context.

Our approach for dialogue-contextualized object detection extends a state-of-the-art object detection model, namely DETR~\cite{carion2020detr}, by injecting dialogue information as the context to guide the detection model to filter out unidentified objects. A similar solution has been proposed by Modulated DETR (MDETR)~\cite{aishwarya2021mdetr}. Despite its strong performance on text-contextualized object detection, MDETR requires an aligned annotation between the text phrase and the visual object for training. Such annotation is not available on SIMMC 2.1, hence we develop a new text-contextualized object detection model namely Situational Context for Multimodal DETR (\textbf{SitCoM-DETR}). Unlike MDETR which concatenates the textual representation along with the visual representation before feeding them into the transformer encoder of DETR (shown in Appendix~\ref{sec:appendix}), \textbf{SitCoM-DETR} injects a dialogue-level semantic representation vector into the input query of the transformer decoder of DETR in order to guide the model to select objects that match the dialogue context. We incorporate the same loss functions as the original DETR model. The depiction of our \textbf{SitCoM-DETR} model is shown in Figure~\ref{fig:sitcom-detr}.

\subsection{Approach 2: Object-Dialogue Alignment}
\label{sec:obj-dialogue-align}

For object-dialogue alignment, we frame the task of multimodal object identification as the alignment between a target object $O^{match}_i$ and a user dialogue turn $D^{user}_i$ pair.
Given a user dialogue turn $D^{user}_i$ and its corresponding scene image $I^{scene}_i$, we first preprocess $I^{scene}_i$ to extract the object images of $O^{match}$.
Each of the object images is paired with $D^{user}_i$ as the positive pairs. We obtain the visual embeddings from the image by feeding it to an image encoder, and the textual embeddings from the dialogue turn by feeding it to a text encoder. After these embeddings pass through a linear projection, we calculate the similarity using the dot product between the two resulting vectors. Utilizing the contrastive learning objective, on a batch of object-dialogue pairs, this cross-modal alignment architecture learns by maximizing the similarity of the positive pairs and minimizing the similarity of the negative pairs (Figure~\ref{fig:contrastive-learning}).


\paragraph{Object-Dialogue Similarity Learning Strategy}

The original contrastive learning approaches the object-dialogue alignment task as a one-to-one function, where the positive sample of $D_i$ is only $O_i$ in Figure~\ref{fig:contrastive-learning}. This is different from the actual nature of multimodal object identification, where more than one object could be relevant to a dialogue turn. For this reason, in addition to the original contrastive learning, we explore two modifications of the learning objective, where: 1) the positive samples of $D_i$ include $O_i$ (image pair) and similar objects\footnote{We define similar objects to $O_i$ as any other objects in the corresponding scene that use the same prefabricated design as $O_i$ in the SIMMC 2.1 dataset.} to $O_i$; and 2) the positive samples of $D_i$ include $O_i$ and other supposedly identified objects in $D_i$.
For simplicity, we refer to these methods as \textbf{CLIPPER (v1)} and \textbf{CLIPPER (v2)}.

\begin{figure*}[t!]
    \centering
    \includegraphics[width=0.85\linewidth, trim={0 0 0 0}, clip]{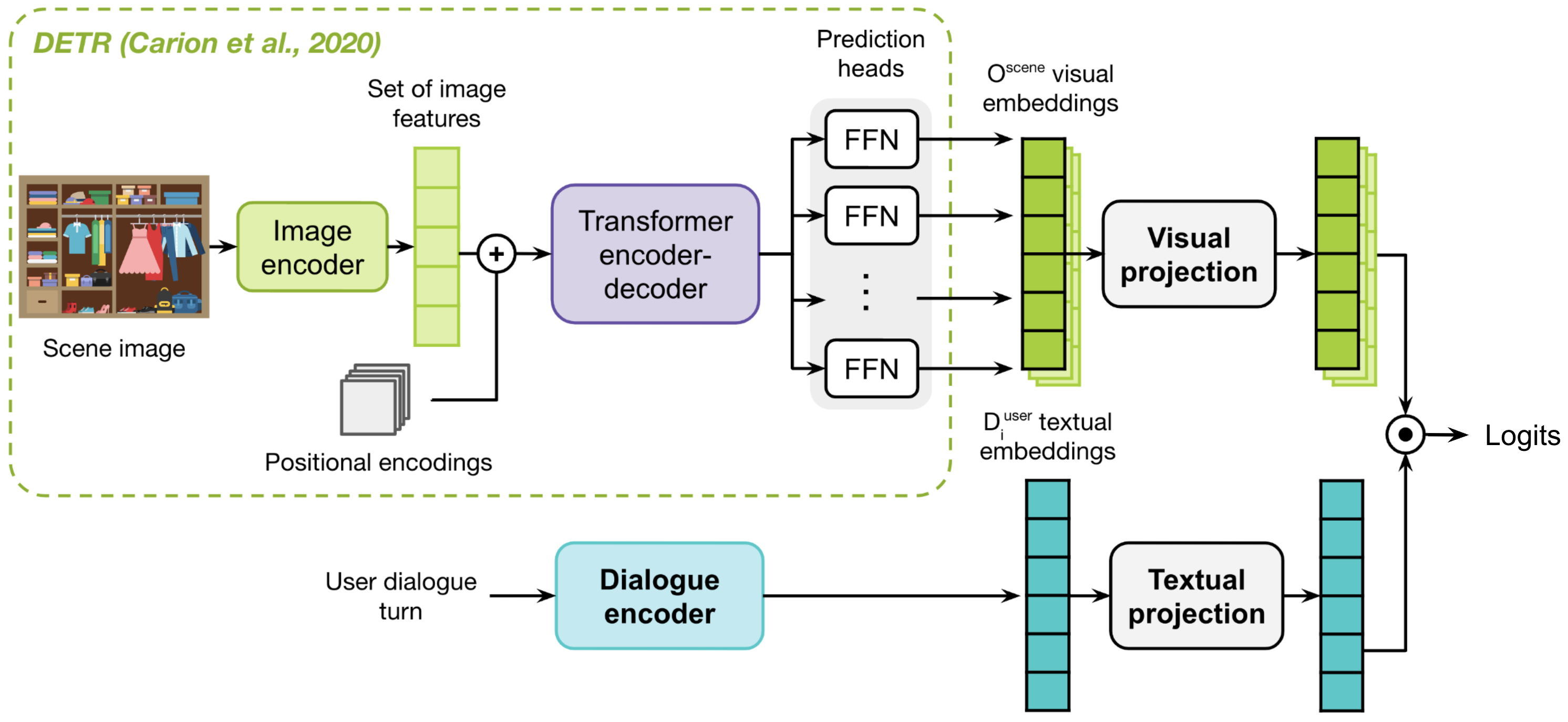}
    \caption{Scene-dialogue alignment. We pre-extract the visual embeddings from an object detection model trained on our dataset. The visual embeddings are used together with dialogue embeddings in the next training to perform multimodal object detection as a binary classification task.}
    \label{fig:scene-dialogue-alignment}
\end{figure*}

\subsection{Approach 3: Scene-Dialogue Alignment}
\label{sec:scene-dialogue-align}

For scene-dialogue alignment, we aim to combine the spatial understanding learned from object detection training with the image-text matching for multimodal similarity learning to solve multimodal object identification. For this approach, we utilize a pre-trained object detection model, i.e., DETR, and two pre-trained language models, i.e., BERT and GPT2. The resulting models are referred to as \textbf{DETR-BERT} and \textbf{DETR-GPT2}, respectively. We illustrate the overview of this approach in Figure~\ref{fig:scene-dialogue-alignment}.

In this approach, we first frame our dataset as an object detection task, where a data instance consists of a scene image $I^{scene}_i$ and its object annotations $O^{scene} = \{(b_1, c_1), (b_2, c_2), ..., (b_m, c_m)\}$, and train an object detection model (DETR) on it. The resulting model is then used to extract the visual representations of all objects in the scene image $I^{scene}$ by matching the object queries with $O^{scene}$ using Hungarian matching~\cite{russell2016hungarianmatcher}.

For the next step, we frame our dataset as a binary classification task, where a data instance consists of a user dialogue turn $D^{user}_i$, an object ${O^{scene}_j}$ in a corresponding scene $I^{scene}_i$, and a binary label (i.e., whether the object is identified by the user dialogue turn or not). We utilize a dialogue encoder to extract textual representation from a user dialogue turn $D^{user}_i$. The textual representation of ${D^{user}_i}$ and the visual representation of ${O^{scene}_j}$ are projected into a latent space. We compute the dot product of the two and use the resulting vector as the prediction logits for training and inference.



\section{Experiment}

\subsection{Dataset}

For all of our experiments, we utilize the ambiguous candidate identification task from the SIMMC 2.1 dataset~\cite{kottur2021simmc}. The dataset studies conversational scenarios where the system shares a co-observed vision (i.e., the same scene) with the user. The dataset focuses on improving the shopping experience in two domains: fashion and furniture. In the setting of SIMMC 2.1, the system is able to access the ground truth meta information of all objects (e.g., object price, size, material, brand, etc.) in the scene $O^{scene}$, while the user observes objects only through the scene viewpoints $\{I^{scene}_1, I^{scene}_2, \dots, I^{scene}_n\}$ to describe a request.

Each dialogue in the dataset can utilize different scene viewpoints at different dialogue turns throughout the session. This represents scenarios where the user navigates the scene during the interaction in a real physical store. Therefore, the multimodal dialogue system needs to understand user requests using both the dialogue history and the scene image as a unified multimodal context. The statistics of the ambiguous candidate identification of SIMMC 2.1 dataset is presented in Table~\ref{tab:dataset-statistics}.\footnote{We use the \texttt{devtest} split of SIMMC 2.1 dataset as the test set in our experiment.}

\subsection{Baselines}

\begin{table}[t!]
    \centering
    \resizebox{0.95\linewidth}{!}{
    \begin{tabular}{cccc}
        \toprule
        \textbf{Split} & \textbf{\# Sample} & \textbf{\# Dialogue} & \textbf{$\boldsymbol{\frac{O^{match}}{O^{scene}}}$} \\
        \toprule
        Train & 4239 & 3983 & 28.74\%\\
        Validation & 414 & 371 & 24.72\%\\
        Test & 940 & 905 & 30.78\%\\
        \bottomrule
    \end{tabular}}
    \caption{Statistics of the ambiguous candidates identification of the SIMMC 2.1 dataset.}
    \label{tab:dataset-statistics}
\end{table}

We incorporate various baselines including simple heuristics and deep learning based multimodal matching methods from SIMMC 2.1.\footnote{SIMMC 2.1 repository: \url{https://github.com/facebookresearch/simmc2}.} For the heuristic methods, we incorporate uniform random prediction (\textbf{Random}), empty prediction (\textbf{No object}), and all objects prediction (\textbf{All objects}) as our baselines. For the deep learning approaches (\textbf{ResNet50-BERT} and \textbf{ResNet50-GPT2}), we apply cosine similarity between the feature extracted from ResNet-50~\cite{he2016resnet}\footnote{We use the pre-extracted visual feature provided in the SIMMC 2.1 repository.} and two widely-used pre-trained LMs, i.e., BERT~\cite{devlin2019bert}\footnote{~\url{https://huggingface.co/bert-base-uncased}.} and GPT2~\cite{radford2019language}\footnote{~\url{https://huggingface.co/gpt2}.}.

In addition to these baselines, we incorporate several additional baselines: 1) pre-trained CLIP~\cite{radford2021clip}\footnote{We use the checkpoint from \url{https://huggingface.co/openai/clip-vit-base-patch32}.}, which serves as a baseline for the object-dialogue alignment approach and 2) pre-trained MDETR~\cite{aishwarya2021mdetr}\footnote{We use the EfficientNet B5 (ENB5) backbone checkpoint from \url{https://github.com/ashkamath/mdetr}.}, which represents a text-conditioned object detection baseline trained with an explicit alignment between phrases and objects. For CLIP, we report both zero-shot (\textbf{CLIP (zero-shot)}) and direct fine-tuning (\textbf{CLIP}) performances, while for MDETR, we only use the zero-shot performance (\textbf{MDETR (zero-shot)}) due to the unavailability of the explicit alignment between objects and dialogues in the dataset.

\subsection{Models}

We propose three different approaches to solve the multimodal object identification task \S\ref{sec:methods}. For the dialogue-contextualized object detection approach, we incorporate one model, namely \textbf{SitCoM-DETR} which will be compared to the MDETR baseline. For the object-dialogue alignment approach, we incorporate two model variants, i.e., \textbf{CLIPPER (v1)} and \textbf{CLIPPER (v2)}.
For the scene-object alignment approach, we incorporate two model variants, i.e., \textbf{DETR-BERT} and \textbf{DETR-GPT2}.

\subsection{Evaluation}

Given a label set $L$ and a prediction set $P$, we define the number of true positive $N^{correct}$ as the objects that appear in both the prediction and the label sets. Using this definition, we evaluate the models' performance on the multimodal object identification task using three evaluation metrics, i.e., recall, precision, and F1-score. The definition of each metric is defined as:
\begin{align}
    &Recall = \frac{N^{correct}}{\|L\|} \\
    &Precision = \frac{N^{correct}}{\|P\|} \\
    &F1 = \frac{2 * Precision * Recall}{Precision + Recall}
\end{align}

\subsection{Implementation Details}
\label{sec:implementation-detail}

\paragraph{Dialogue Preprocessing}
In all of our experiments, following prior works in end-to-end task-oriented dialogue system, we encode the last three utterances from the dialogue into a single text. For example a user dialogue turn $D^{user}_i = \{u_1, s_1, u_2, s_2, \dots, u_i\}$ is encoded into a text "U: <$u_{i-1}$> S: <$s_{i-1}$> U: <$u_i$>" to be further processed by the dialogue encoder.

\paragraph{Inference strategy for object-dialogue alignment}
For the proposed CLIPPER model in the object-dialogue alignment approach, we simply apply sigmoid to the logits and use a threshold value of 0.5 (denoted as \textit{Sigmoid}), since it has a built-in capability to perform multi-label classification. While for the CLIP model, which serves as a baseline, does not have the same capability, hence we use the mean value of the logits as the threshold (denoted as \textit{Mean}). Additionally, we also evaluate the performance of the model if the top-$k$ objects with the highest logits are considered valid predictions, where $k$ denotes the correct amount of objects in the ground-truth label (denoted as \textit{Oracle}).

\paragraph{Inference strategy for dialogue-contextualized object detection}
For the dialogue-contextualized object detection, since the model is originally for the object detection task, we develop our own inference strategy to allow it to perform multi-label classification for object identification. This is done through several steps: 1) we perform Hungarian matching using all objects, 2) we compute intersection over union (IoU) of all pairs of matched prediction and ground-truth bounding boxes\footnote{We do not consider the class label in the scoring to have a fairer comparison with the zero-shot MDETR approach.}, and 3) we take all objects having IoU score $\geq$10\%\footnote{We align this with MDETR's class probability setting during inference.}.

\begin{table*}[t!]
    \centering
    \resizebox{0.925\linewidth}{!}{
        \begin{tabular}{p{4cm}lccc}
            \toprule
             \textbf{Method Type} & \textbf{Approach} & \textbf{Recall} & \textbf{Precision} & \textbf{F1-score} \\
            \toprule
            \multicolumn{5}{>{\columncolor[gray]{.9}}c}{\textbf{\textit{Baselines}}} \\
            \multirow{3}{*}{\textit{Heuristic}} & No object & 0.00\% & 0.00\% & 0.00\% \\
             & Random & 49.90\% & \underline{22.43\%} & 30.95\% \\
             & All objects & \textbf{100.00}\% & 22.34\% & \underline{36.52\%} \\
            \midrule
            \multirow{2}{4cm}{\textit{SIMMC 2.1}} & ResNet50-GPT2 & 36.40\% & 42.26\% & 39.11\% \\
             & ResNet50-BERT & \underline{36.70\%} & \textbf{43.39\%} & \underline{39.76\%} \\
            \midrule
            \multirow{2}{4cm}{\textit{Dialogue-Contextualized Object Detection}} & \multirow{2}{*}{MDETR (zero-shot)} & \multirow{2}{*}{\underline{16.33\%}} & \multirow{2}{*}{\underline{29.70\%}} & \multirow{2}{*}{\underline{21.07\%}} \\
            & & & & \\
            \midrule
            \multirow{2}{4cm}{\textit{Object-Dialogue Alignment}} & CLIP (zero-shot) & 55.70\% & 26.39\% & 35.81\% \\
             & CLIP (fine-tuned) & \underline{73.00\%} & \underline{32.62\%} & \textbf{45.09\%} \\
            \toprule
            \multicolumn{5}{>{\columncolor[gray]{.9}}c}{\textbf{\textit{Proposed Methods}}} \\
            \multirow{2}{4cm}{\textit{Dialogue-Contextualized Object Detection}} & SitCoM-DETR (aug) & 47.82\% & 25.69\% & 33.42\% \\
             & SitCoM-DETR (no aug) & \underline{49.51\%} & \underline{25.81\%} & \underline{33.93\%} \\
            \midrule
            \multirow{2}{4cm}{\textit{Object-Dialogue Alignment}} & CLIPPER (v1) & \textbf{73.41\%} & \underline{33.00\%} & \underline{45.53\%} \\
             & CLIPPER (v2) & 59.95\% & 25.60\% & 35.88\% \\
            \midrule
            \multirow{2}{4cm}{\textit{Scene-Dialogue Alignment}} & DETR-BERT & \underline{65.47\%} & 51.48\% & 57.64\% \\
             & DETR-GPT2 & 63.81\% & \textbf{56.79\%} & \textbf{60.10\%} \\
            \bottomrule
        \end{tabular}
    }
    \caption{Experimental results of multimodal object identification on the SIMMC 2.1 dataset~\cite{kottur2021simmc}. \textbf{Bold} denotes the best performances of baselines and proposed methods. \underline{Underline} denotes the best performances within a method type.}
    \label{tab:results}
\end{table*}

\paragraph{Hyperparameter Details}
For the dialogue-contextualized object detection, we fine-tune the SitCoM-DETR model for a maximum of 200 epochs with AdamW optimizer using a linear learning rate decay, a learning rate between [1e-4..1e-5], and an early stopping of 10 epochs. 
For the object-dialogue alignment, we fine-tune the CLIP and CLIPPER models for a maximum of 200 epochs with AdamW optimizer using a linear learning rate decay, a learning rate between [1e-4..1e-5], and an early stopping of 10 epochs. 
For the scene-dialogue alignment, we fine-tune the DETR-BERT and DETR-GPT2 models for a maximum of 200 epochs with AdamW optimizer using a linear learning rate decay, a learning rate between [1e-4..1e-5], and an early stopping of 10 epochs.

\section{Result and Analysis}
\label{sec:result}

\subsection{Result Overview}

The results of our experiments are shown in Table~\ref{tab:results}. The best baseline performance is achieved by \textbf{CLIP (fine-tuned)} with 45.09\% F1-score outperforming the baselines provided by the SIMMC 2.1 (i.e., \textbf{ResNet50-GPT2} and \textbf{ResNet50-BERT}), showing the superiority of image-text alignment pre-training over separate unimodal pre-trainings for multimodal object identification. For the dialogue-contextualized object detection methods, the proposed \textbf{SitCoM-DETR} outperforms \textbf{MDETR (zero-shot)}. Nevertheless, its performance for multimodal object identification is low despite having an acceptable object detection quality.
We conjecture that a better method for adapting an object detection model for multimodal object identification is required, which is also shown by our \textit{scene-dialogue alignment} approach in \S\ref{sec:scene-dialogue-align}.

For the object-dialogue alignment, our \textbf{CLIPPER (v1)} marginally outperforms the \textbf{CLIP (fine-tuned)} baseline. This shows the effectiveness of modifying the CLIP objective which is explained in more detail in \S\ref{sec:clip-objective}. For the scene-dialogue alignment (i.e., \textbf{DETR-BERT} and \textbf{DETR-GPT2}), where we combine the object detection and the image-text contrastive objective, we show a significant improvement over \textbf{CLIP (fine-tuned)}, which is the highest-performing baseline, by $\sim$10-15\% F1-score. This suggests the importance of combining object detection representation and image-text contrastive learning to fulfill the need for both visual and spatial matching to solve multimodal object identification.

\subsection{Pitfalls of the Best Performing Models}
\label{sec:error-analysis}

\begin{figure}
    \centering
    \resizebox{1\linewidth}{!}{
        \includegraphics{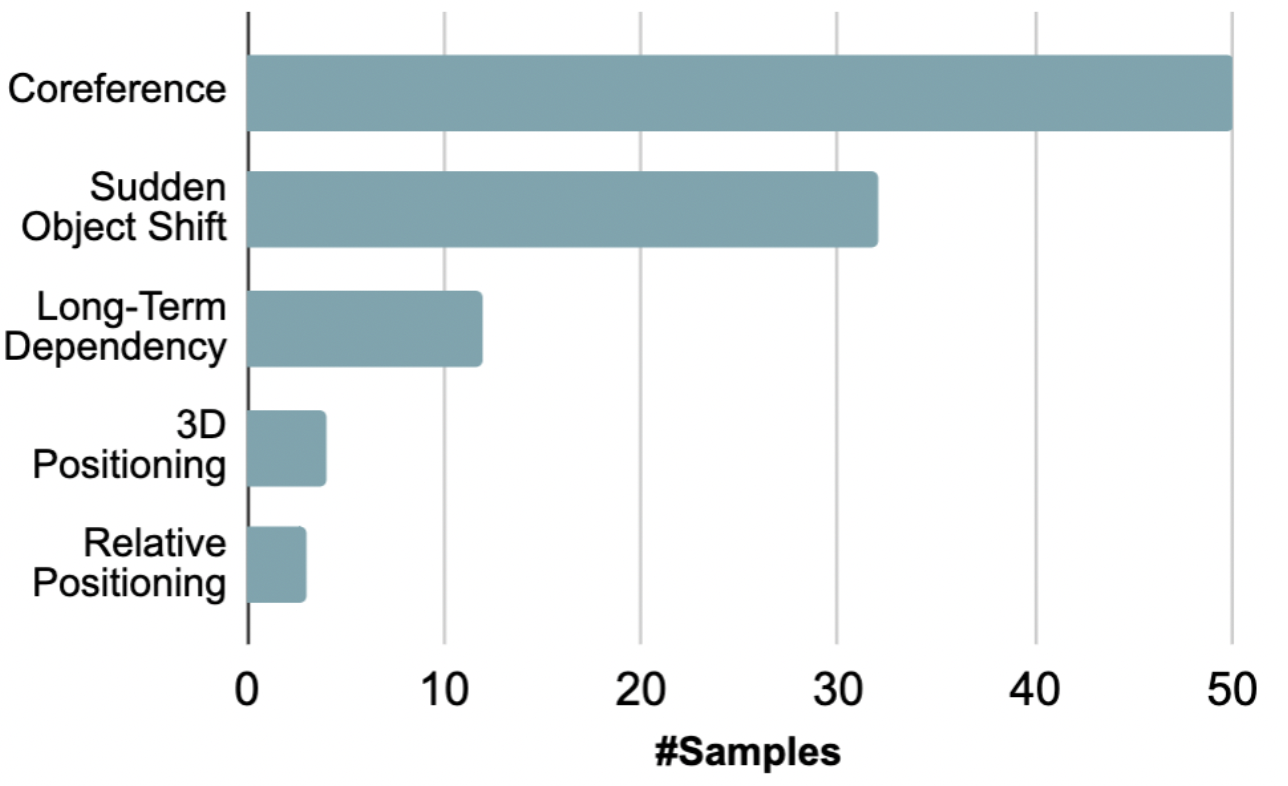}
    }
    \caption{Frequency of error types of 100 misclassified samples from \textbf{DETR-BERT} and \textbf{DETR-GPT2}.}
    \label{fig:error}
\end{figure}

We manually analyze the incorrect predictions made by our scene-dialogue alignment approaches, i.e., \textbf{DETR-BERT} and \textbf{DETR-GPT2}.
Based on our analysis in Table~\ref{fig:error}, our models encounter two main issues. First, our models have difficulties in identifying objects when faced with a sudden object shift in the dialogue, e.g., the sudden shift from beds to a chair in this user dialogue turn U: ``\textit{I need \ul{a new bed} too. Any suggestions?}'', S: ``\textit{Both of \ul{these grey beds} are in stock.}'', U: ``\textit{What's the rating on \ul{that chair}?}''.

The second issue is the ineffectiveness of handling textual coreferences. For instance, in the user dialogue turn U: ``\textit{How about a hat, but cheap and in a small?}'', S: ``\textit{I have \ul{the black hat third from the front}, the white hat at the front, and \ul{the black hat between them}.}'', U: ``\textit{What's the brand and reviews for \ul{the black hat}?}'', the models fail to recognize that ``\ul{the black hat}'' in the user utterance is anaphoric to either ``\ul{the black hat third from the front}'' or ``\ul{the black hat between them}'' in the system utterance, which leads to the system's failure to identify both black hats as $O^{match}$. This shortcoming also becomes more pronounced if the coreference chains are longer.


These issues show the limitation of pre-trained LMs for discourse understanding and analysis, especially in terms of coreference and entity linking~\cite{jurafsky2019speech, pandia-etal-2021-pragmatic, koto-etal-2021-discourse}. Additionally, some other cases require the model to process long-term dialogue history dependency which existing LMs are not able to handle because of the quadratic cost bottleneck of the attention mechanism of the transformer architecture~\cite{vaswani2017attention}. Adapting an efficient attention mechanism with linear complexity might be beneficial to mitigate this problem.

\subsection{Impact of Changing CLIP Objective}
\label{sec:clip-objective}

As shown in Table~\ref{tab:object-dialogue-analysis}, the CLIPPER models with binary cross-entropy objective have a built-in capability for multi-label classification with \textbf{Sigmoid} which consistently performs better compared to the \textbf{Mean} thresholding. In addition, \textbf{CLIPPER (v1)} outperforms the original CLIP model which is trained with the cross-entropy loss. These facts suggest that changing the CLIP objective is beneficial for performing multi-label classification tasks such as multimodal object identification.

When using \textbf{Oracle}, we can observe a significant improvement in F1-score score, which mainly comes from the improvement in the precision with only a minor degradation on recall. This suggests that there is a very sensitive range of logits which consists of many negative samples with a few positive samples. To better segregate these few positive samples from the negative ones, hard negative mining techniques such as focal loss~\cite{lin2020focal} might be beneficial to alleviate this problem.

\begin{table}[t!]
    \centering
    \resizebox{0.95\linewidth}{!}{
    \begin{tabular}{lccc}
        \toprule
         \textbf{Approach} & \textbf{Rec.} & \textbf{Prec.} & \textbf{F1} \\
        \toprule
        \multicolumn{4}{l}{\textbf{CLIP --- Cross-Entropy}} \\
        \hspace{8mm} Mean & 73.00\% & 32.62\% & \underline{45.09\%} \\
        \hspace{8mm} Oracle & 74.99\% & 74.96\% & \textbf{74.98\%} \\
        \midrule
        \multicolumn{4}{l}{\textbf{CLIPPER (v1) --- Binary Cross-Entropy}} \\
        \hspace{8mm} Sigmoid & 73.41\% & 33.00\% & \underline{45.53\%} \\
        \hspace{8mm} Mean & 73.08\% & 31.97\% & 44.48\% \\
        \hspace{8mm} Oracle & 73.37\% & 73.34\% & \textbf{73.36\%} \\
        \midrule
        \multicolumn{4}{l}{\textbf{CLIPPER (v2) --- Binary Cross-Entropy}} \\
        \hspace{8mm} Sigmoid & 59.95\% & 25.60\% & \underline{35.88\%} \\
        \hspace{8mm} Mean & 53.90\% & 23.42\% & 32.65\% \\
        \hspace{8mm} Oracle & 54.92\% & 54.89\% & \textbf{54.91\%} \\
        \bottomrule
    \end{tabular}}
    \caption{Results for object-dialogue alignment models with different thresholding strategies.}
    \label{tab:object-dialogue-analysis}
\end{table}

\section{Discussion}

Based on the results and analysis, we show that the \textit{scene-object alignment} approach is the best performing approach, achieving $\sim$55-60\% F1-score in the multimodal object identification task of SIMMC 2.1. We analyze the behavior of the model and conjecture that existing LMs have a limitation on understanding discourse. Additionally, we show the potential benefit of modeling the long-term dependency of dialogue history to further improve the quality of multimodal object identification task (\S\ref{sec:error-analysis}). Lastly, we analyze the limitation of the existing image-text contrastive approaches for multimodal object identification and propose an alternative objective to alleviate this limitation (\S\ref{sec:clip-objective}).

For future work, we aim to focus on the scene-dialogue alignment methods to further improve the model performance on the multimodal object identification capability. We note five potential points of improvement that can be further explored to improve the model performance in multimodal object identification: 1) the incorporation of cross-object attention in the modality fusion phase to enable a better relative position understanding between objects, 2) the incorporation of linear attention mechanism to handle the long-term dependency of dialogue history, 3) the exploration on better contrastive objectives for multimodal object identification, 4) the exploration on improving discourse understanding for LMs to better handle coreference and sudden object shift, and 5) the synthetic scene-dialogue data augmentation through the utilization of other publicly available object detection datasets to handle the in-domain data scarcity problem.

\section{Conclusion}
In this paper, we explore three methods to tackle multimodal object identification and evaluate them on SIMMC 2.1. Our best method, scene-dialogue alignment, improves the performance by $\sim$20\% F1-score compared to the SIMMC 2.1 baselines. We provide an analysis of incorrect predictions by our best approach and the impact of changing the CLIP learning objective. We further provide discussion regarding the limitation of our methods and the potential directions for future works.

\section*{Acknowledgement}

We appreciate the guidance that Prof. Dan Xu has provided for this research.
This work has been supported by the School of Engineering PhD Fellowship Award, the Hong Kong University of Science and Technology and PF20-43679 Hong Kong PhD Fellowship Scheme, Research Grant Council, Hong Kong.

\bibliography{anthology,custom}

\begin{thebibliography}{50}
\expandafter\ifx\csname natexlab\endcsname\relax\def\natexlab#1{#1}\fi

\bibitem[{Alamri et~al.(2019)Alamri, Cartillier, Das, Wang, Cherian, Essa,
  Batra, Marks, Hori, Anderson et~al.}]{alamri2019audio}
Huda Alamri, Vincent Cartillier, Abhishek Das, Jue Wang, Anoop Cherian, Irfan
  Essa, Dhruv Batra, Tim~K Marks, Chiori Hori, Peter Anderson, et~al. 2019.
\newblock Audio visual scene-aware dialog.
\newblock In \emph{Proceedings of the IEEE/CVF Conference on Computer Vision
  and Pattern Recognition}, pages 7558--7567.

\bibitem[{Antol et~al.(2015)Antol, Agrawal, Lu, Mitchell, Batra, Zitnick, and
  Parikh}]{antol2015vqa}
Stanislaw Antol, Aishwarya Agrawal, Jiasen Lu, Margaret Mitchell, Dhruv Batra,
  C~Lawrence Zitnick, and Devi Parikh. 2015.
\newblock Vqa: Visual question answering.
\newblock In \emph{Proceedings of the IEEE international conference on computer
  vision}, pages 2425--2433.

\bibitem[{Brawer et~al.(2018)Brawer, Mangin, Roncone, Widder, and
  Scassellati}]{brawer2018situated}
Jake Brawer, Olivier Mangin, Alessandro Roncone, Sarah Widder, and Brian
  Scassellati. 2018.
\newblock Situated human--robot collaboration: predicting intent from grounded
  natural language.
\newblock In \emph{2018 IEEE/RSJ International Conference on Intelligent Robots
  and Systems (IROS)}, pages 827--833. IEEE.

\bibitem[{Carion et~al.(2020)Carion, Massa, Synnaeve, Usunier, Kirillov, and
  Zagoruyko}]{carion2020detr}
Nicolas Carion, Francisco Massa, Gabriel Synnaeve, Nicolas Usunier, Alexander
  Kirillov, and Sergey Zagoruyko. 2020.
\newblock End-to-end object detection with transformers.
\newblock In \emph{Computer Vision -- ECCV 2020}, pages 213--229, Cham.
  Springer International Publishing.

\bibitem[{Dai et~al.(2021)Dai, Cahyawijaya, Liu, and
  Fung}]{dai-etal-2021-multimodal}
Wenliang Dai, Samuel Cahyawijaya, Zihan Liu, and Pascale Fung. 2021.
\newblock \href {https://doi.org/10.18653/v1/2021.naacl-main.417} {Multimodal
  end-to-end sparse model for emotion recognition}.
\newblock In \emph{Proceedings of the 2021 Conference of the North American
  Chapter of the Association for Computational Linguistics: Human Language
  Technologies}, pages 5305--5316, Online. Association for Computational
  Linguistics.

\bibitem[{Das et~al.(2017)Das, Kottur, Gupta, Singh, Yadav, Moura, Parikh, and
  Batra}]{das2017visual}
Abhishek Das, Satwik Kottur, Khushi Gupta, Avi Singh, Deshraj Yadav,
  Jos{\'e}~MF Moura, Devi Parikh, and Dhruv Batra. 2017.
\newblock Visual dialog.
\newblock In \emph{Proceedings of the IEEE conference on computer vision and
  pattern recognition}, pages 326--335.

\bibitem[{Devlin et~al.(2019)Devlin, Chang, Lee, and
  Toutanova}]{devlin2019bert}
Jacob Devlin, Ming-Wei Chang, Kenton Lee, and Kristina Toutanova. 2019.
\newblock \href {https://doi.org/10.18653/v1/N19-1423} {{BERT}: Pre-training of
  deep bidirectional transformers for language understanding}.
\newblock In \emph{Proceedings of the 2019 Conference of the North {A}merican
  Chapter of the Association for Computational Linguistics: Human Language
  Technologies, Volume 1 (Long and Short Papers)}, pages 4171--4186,
  Minneapolis, Minnesota. Association for Computational Linguistics.

\bibitem[{Gkatzia et~al.(2015)Gkatzia, Rieser, Bartie, and
  Mackaness}]{gkatzia-etal-2015-virtual}
Dimitra Gkatzia, Verena Rieser, Phil Bartie, and William Mackaness. 2015.
\newblock \href {https://doi.org/10.18653/v1/D15-1224} {From the virtual to the
  {R}eal{W}orld: Referring to objects in real-world spatial scenes}.
\newblock In \emph{Proceedings of the 2015 Conference on Empirical Methods in
  Natural Language Processing}, pages 1936--1942, Lisbon, Portugal. Association
  for Computational Linguistics.

\bibitem[{He et~al.(2016)He, Zhang, Ren, and Sun}]{he2016resnet}
Kaiming He, Xiangyu Zhang, Shaoqing Ren, and Jian Sun. 2016.
\newblock \href {https://doi.org/10.1109/CVPR.2016.90} {Deep residual learning
  for image recognition}.
\newblock In \emph{2016 IEEE Conference on Computer Vision and Pattern
  Recognition (CVPR)}, pages 770--778.

\bibitem[{Hossain et~al.(2019)Hossain, Sohel, Shiratuddin, and
  Laga}]{hossain2019comprehensive}
MD~Zakir Hossain, Ferdous Sohel, Mohd~Fairuz Shiratuddin, and Hamid Laga. 2019.
\newblock A comprehensive survey of deep learning for image captioning.
\newblock \emph{ACM Computing Surveys (CsUR)}, 51(6):1--36.

\bibitem[{Hu et~al.(2016)Hu, Xu, Rohrbach, Feng, Saenko, and
  Darrell}]{hu2016natural}
Ronghang Hu, Huazhe Xu, Marcus Rohrbach, Jiashi Feng, Kate Saenko, and Trevor
  Darrell. 2016.
\newblock Natural language object retrieval.
\newblock In \emph{Proceedings of the IEEE conference on computer vision and
  pattern recognition}, pages 4555--4564.

\bibitem[{Ilinykh et~al.(2019)Ilinykh, Zarrie{\ss}, and
  Schlangen}]{ilinykh-etal-2019-tell}
Nikolai Ilinykh, Sina Zarrie{\ss}, and David Schlangen. 2019.
\newblock \href {https://doi.org/10.18653/v1/W19-8621} {Tell me more: A dataset
  of visual scene description sequences}.
\newblock In \emph{Proceedings of the 12th International Conference on Natural
  Language Generation}, pages 152--157, Tokyo, Japan. Association for
  Computational Linguistics.

\bibitem[{Jia et~al.(2021)Jia, Yang, Xia, Chen, Parekh, Pham, Le, Sung, Li, and
  Duerig}]{jia2021scaling}
Chao Jia, Yinfei Yang, Ye~Xia, Yi-Ting Chen, Zarana Parekh, Hieu Pham, Quoc Le,
  Yun-Hsuan Sung, Zhen Li, and Tom Duerig. 2021.
\newblock Scaling up visual and vision-language representation learning with
  noisy text supervision.
\newblock In \emph{International Conference on Machine Learning}, pages
  4904--4916. PMLR.

\bibitem[{Johnson et~al.(2017)Johnson, Hariharan, Van Der~Maaten, Fei-Fei,
  Lawrence~Zitnick, and Girshick}]{johnson2017clevr}
Justin Johnson, Bharath Hariharan, Laurens Van Der~Maaten, Li~Fei-Fei,
  C~Lawrence~Zitnick, and Ross Girshick. 2017.
\newblock Clevr: A diagnostic dataset for compositional language and elementary
  visual reasoning.
\newblock In \emph{Proceedings of the IEEE conference on computer vision and
  pattern recognition}, pages 2901--2910.

\bibitem[{Jurafsky and Martin(2019)}]{jurafsky2019speech}
Dan Jurafsky and James~H Martin. 2019.
\newblock \emph{Speech and Language Processing: An Introduction to Natural
  Language Processing, Computational Linguistics, and Speech Recognition (3rd
  draft ed.)}.
\newblock Stanford Univ.

\bibitem[{Kamath et~al.(2021)Kamath, Singh, LeCun, Synnaeve, Misra, and
  Carion}]{aishwarya2021mdetr}
Aishwarya Kamath, Mannat Singh, Yann LeCun, Gabriel Synnaeve, Ishan Misra, and
  Nicolas Carion. 2021.
\newblock \href {https://doi.org/10.1109/iccv48922.2021.00180} {{MDETR} -
  modulated detection for end-to-end multi-modal understanding}.
\newblock In \emph{2021 {IEEE}/{CVF} International Conference on Computer
  Vision ({ICCV})}. {IEEE}.

\bibitem[{Kazemzadeh et~al.(2014)Kazemzadeh, Ordonez, Matten, and
  Berg}]{kazemzadeh-etal-2014-referitgame}
Sahar Kazemzadeh, Vicente Ordonez, Mark Matten, and Tamara Berg. 2014.
\newblock \href {https://doi.org/10.3115/v1/D14-1086} {{R}efer{I}t{G}ame:
  Referring to objects in photographs of natural scenes}.
\newblock In \emph{Proceedings of the 2014 Conference on Empirical Methods in
  Natural Language Processing ({EMNLP})}, pages 787--798, Doha, Qatar.
  Association for Computational Linguistics.

\bibitem[{Koto et~al.(2021)Koto, Lau, and Baldwin}]{koto-etal-2021-discourse}
Fajri Koto, Jey~Han Lau, and Timothy Baldwin. 2021.
\newblock \href {https://doi.org/10.18653/v1/2021.naacl-main.301} {Discourse
  probing of pretrained language models}.
\newblock In \emph{Proceedings of the 2021 Conference of the North American
  Chapter of the Association for Computational Linguistics: Human Language
  Technologies}, pages 3849--3864, Online. Association for Computational
  Linguistics.

\bibitem[{Kottur et~al.(2021)Kottur, Moon, Geramifard, and
  Damavandi}]{kottur2021simmc}
Satwik Kottur, Seungwhan Moon, Alborz Geramifard, and Babak Damavandi. 2021.
\newblock \href {https://doi.org/10.18653/v1/2021.emnlp-main.401} {{SIMMC} 2.0:
  A task-oriented dialog dataset for immersive multimodal conversations}.
\newblock In \emph{Proceedings of the 2021 Conference on Empirical Methods in
  Natural Language Processing}, pages 4903--4912, Online and Punta Cana,
  Dominican Republic. Association for Computational Linguistics.

\bibitem[{Kottur et~al.(2019)Kottur, Moura, Parikh, Batra, and
  Rohrbach}]{kottur-etal-2019-clevr}
Satwik Kottur, Jos{\'e} M.~F. Moura, Devi Parikh, Dhruv Batra, and Marcus
  Rohrbach. 2019.
\newblock \href {https://doi.org/10.18653/v1/N19-1058} {{CLEVR}-dialog: A
  diagnostic dataset for multi-round reasoning in visual dialog}.
\newblock In \emph{Proceedings of the 2019 Conference of the North {A}merican
  Chapter of the Association for Computational Linguistics: Human Language
  Technologies, Volume 1 (Long and Short Papers)}, pages 582--595, Minneapolis,
  Minnesota. Association for Computational Linguistics.

\bibitem[{Kottur et~al.(2018)Kottur, Moura, Parikh, Batra, and
  Rohrbach}]{kottur2018visual}
Satwik Kottur, Jos{\'e}~MF Moura, Devi Parikh, Dhruv Batra, and Marcus
  Rohrbach. 2018.
\newblock Visual coreference resolution in visual dialog using neural module
  networks.
\newblock In \emph{Proceedings of the European Conference on Computer Vision
  (ECCV)}, pages 153--169.

\bibitem[{Kuo and Kira(2022)}]{kuo2022beyond}
Chia-Wen Kuo and Zsolt Kira. 2022.
\newblock Beyond a pre-trained object detector: Cross-modal textual and visual
  context for image captioning.
\newblock In \emph{Proceedings of the IEEE/CVF Conference on Computer Vision
  and Pattern Recognition}, pages 17969--17979.

\bibitem[{Landragin(2006)}]{landragin2006visual}
Fr{\'e}d{\'e}ric Landragin. 2006.
\newblock Visual perception, language and gesture: A model for their
  understanding in multimodal dialogue systems.
\newblock \emph{Signal Processing}, 86(12):3578--3595.

\bibitem[{Li et~al.(2019)Li, Yao, Zhang, Wang, Kanhere, and Zhang}]{li2019zero}
Zhihui Li, Lina Yao, Xiaoqin Zhang, Xianzhi Wang, Salil Kanhere, and Huaxiang
  Zhang. 2019.
\newblock Zero-shot object detection with textual descriptions.
\newblock In \emph{Proceedings of the AAAI Conference on Artificial
  Intelligence}, volume~33, pages 8690--8697.

\bibitem[{Lin et~al.(2020)Lin, Goyal, Girshick, He, and Dollár}]{lin2020focal}
Tsung-Yi Lin, Priya Goyal, Ross Girshick, Kaiming He, and Piotr Dollár. 2020.
\newblock \href {https://doi.org/10.1109/TPAMI.2018.2858826} {Focal loss for
  dense object detection}.
\newblock \emph{IEEE Transactions on Pattern Analysis and Machine
  Intelligence}, 42(2):318--327.

\bibitem[{Liu et~al.(2019)Liu, Liu, Bai, and Yuille}]{liu2019clevr}
Runtao Liu, Chenxi Liu, Yutong Bai, and Alan~L Yuille. 2019.
\newblock Clevr-ref+: Diagnosing visual reasoning with referring expressions.
\newblock In \emph{Proceedings of the IEEE/CVF Conference on Computer Vision
  and Pattern Recognition}, pages 4185--4194.

\bibitem[{Lo{\'a}iciga et~al.(2021{\natexlab{a}})Lo{\'a}iciga, Dobnik, and
  Schlangen}]{loaiciga-etal-2021-annotating}
Sharid Lo{\'a}iciga, Simon Dobnik, and David Schlangen. 2021{\natexlab{a}}.
\newblock \href {https://aclanthology.org/2021.mmsr-1.7} {Annotating anaphoric
  phenomena in situated dialogue}.
\newblock In \emph{Proceedings of the 1st Workshop on Multimodal Semantic
  Representations (MMSR)}, pages 78--88, Groningen, Netherlands (Online).
  Association for Computational Linguistics.

\bibitem[{Lo{\'a}iciga et~al.(2021{\natexlab{b}})Lo{\'a}iciga, Dobnik, and
  Schlangen}]{loaiciga-etal-2021-reference}
Sharid Lo{\'a}iciga, Simon Dobnik, and David Schlangen. 2021{\natexlab{b}}.
\newblock \href {https://doi.org/10.18653/v1/2021.alvr-1.7} {Reference and
  coreference in situated dialogue}.
\newblock In \emph{Proceedings of the Second Workshop on Advances in Language
  and Vision Research}, pages 39--44, Online. Association for Computational
  Linguistics.

\bibitem[{Lovenia et~al.(2022)Lovenia, Wilie, Barraud, Cahyawijaya, Chung, and
  Fung}]{lovenia-etal-2022-every}
Holy Lovenia, Bryan Wilie, Romain Barraud, Samuel Cahyawijaya, Willy Chung, and
  Pascale Fung. 2022.
\newblock \href {https://aclanthology.org/2022.latechclfl-1.6} {Every picture
  tells a story: Image-grounded controllable stylistic story generation}.
\newblock In \emph{Proceedings of the 6th Joint SIGHUM Workshop on
  Computational Linguistics for Cultural Heritage, Social Sciences, Humanities
  and Literature}, pages 40--52, Gyeongju, Republic of Korea. International
  Conference on Computational Linguistics.

\bibitem[{Lukin et~al.(2018)Lukin, Gervits, Hayes, Moolchandani, Leuski,
  Rogers~III, Sanchez~Amaro, Marge, Voss, and Traum}]{lukin-etal-2018-scoutbot}
Stephanie~M. Lukin, Felix Gervits, Cory~J. Hayes, Pooja Moolchandani, Anton
  Leuski, John~G. Rogers~III, Carlos Sanchez~Amaro, Matthew Marge, Clare~R.
  Voss, and David Traum. 2018.
\newblock \href {https://doi.org/10.18653/v1/P18-4016} {{S}cout{B}ot: A
  dialogue system for collaborative navigation}.
\newblock In \emph{Proceedings of {ACL} 2018, System Demonstrations}, pages
  93--98, Melbourne, Australia. Association for Computational Linguistics.

\bibitem[{Mao et~al.(2016)Mao, Huang, Toshev, Camburu, Yuille, and
  Murphy}]{mao2016generation}
Junhua Mao, Jonathan Huang, Alexander Toshev, Oana Camburu, Alan~L Yuille, and
  Kevin Murphy. 2016.
\newblock Generation and comprehension of unambiguous object descriptions.
\newblock In \emph{Proceedings of the IEEE conference on computer vision and
  pattern recognition}, pages 11--20.

\bibitem[{Min et~al.(2021)Min, Dang, and Moon}]{min2021deep}
Kyungbok Min, Minh Dang, and Hyeonjoon Moon. 2021.
\newblock Deep learning-based short story generation for an image using the
  encoder-decoder structure.
\newblock \emph{IEEE Access}, 9:113550--113557.

\bibitem[{Mitchell et~al.(2010)Mitchell, van Deemter, and
  Reiter}]{mitchell2010natural}
Margaret Mitchell, Kees van Deemter, and Ehud Reiter. 2010.
\newblock Natural reference to objects in a visual domain.
\newblock In \emph{Proceedings of the 6th international natural language
  generation conference}.

\bibitem[{Moon et~al.(2020)Moon, Kottur, Crook, De, Poddar, Levin, Whitney,
  Difranco, Beirami, Cho, Subba, and Geramifard}]{moon-etal-2020-situated}
Seungwhan Moon, Satwik Kottur, Paul Crook, Ankita De, Shivani Poddar, Theodore
  Levin, David Whitney, Daniel Difranco, Ahmad Beirami, Eunjoon Cho, Rajen
  Subba, and Alborz Geramifard. 2020.
\newblock \href {https://doi.org/10.18653/v1/2020.coling-main.96} {Situated and
  interactive multimodal conversations}.
\newblock In \emph{Proceedings of the 28th International Conference on
  Computational Linguistics}, pages 1103--1121, Barcelona, Spain (Online).
  International Committee on Computational Linguistics.

\bibitem[{Mostafazadeh et~al.(2017)Mostafazadeh, Brockett, Dolan, Galley, Gao,
  Spithourakis, and Vanderwende}]{mostafazadeh-etal-2017-image}
Nasrin Mostafazadeh, Chris Brockett, Bill Dolan, Michel Galley, Jianfeng Gao,
  Georgios Spithourakis, and Lucy Vanderwende. 2017.
\newblock \href {https://aclanthology.org/I17-1047} {Image-grounded
  conversations: Multimodal context for natural question and response
  generation}.
\newblock In \emph{Proceedings of the Eighth International Joint Conference on
  Natural Language Processing (Volume 1: Long Papers)}, pages 462--472, Taipei,
  Taiwan. Asian Federation of Natural Language Processing.

\bibitem[{Pandia et~al.(2021)Pandia, Cong, and
  Ettinger}]{pandia-etal-2021-pragmatic}
Lalchand Pandia, Yan Cong, and Allyson Ettinger. 2021.
\newblock \href {https://doi.org/10.18653/v1/2021.conll-1.29} {Pragmatic
  competence of pre-trained language models through the lens of discourse
  connectives}.
\newblock In \emph{Proceedings of the 25th Conference on Computational Natural
  Language Learning}, pages 367--379, Online. Association for Computational
  Linguistics.

\bibitem[{Radford et~al.(2021)Radford, Kim, Hallacy, Ramesh, Goh, Agarwal,
  Sastry, Askell, Mishkin, Clark, Krueger, and Sutskever}]{radford2021clip}
Alec Radford, Jong~Wook Kim, Chris Hallacy, Aditya Ramesh, Gabriel Goh,
  Sandhini Agarwal, Girish Sastry, Amanda Askell, Pamela Mishkin, Jack Clark,
  Gretchen Krueger, and Ilya Sutskever. 2021.
\newblock \href {https://proceedings.mlr.press/v139/radford21a.html} {Learning
  transferable visual models from natural language supervision}.
\newblock In \emph{Proceedings of the 38th International Conference on Machine
  Learning}, volume 139 of \emph{Proceedings of Machine Learning Research},
  pages 8748--8763. PMLR.

\bibitem[{Radford et~al.(2019)Radford, Wu, Child, Luan, Amodei, and
  Sutskever}]{radford2019language}
Alec Radford, Jeff Wu, Rewon Child, David Luan, Dario Amodei, and Ilya
  Sutskever. 2019.
\newblock Language models are unsupervised multitask learners.

\bibitem[{Saha et~al.(2018)Saha, Khapra, and
  Sankaranarayanan}]{saha2018towards}
Amrita Saha, Mitesh Khapra, and Karthik Sankaranarayanan. 2018.
\newblock Towards building large scale multimodal domain-aware conversation
  systems.
\newblock In \emph{Proceedings of the AAAI Conference on Artificial
  Intelligence}, volume~32.

\bibitem[{Seo et~al.(2017)Seo, Lehrmann, Han, and Sigal}]{seo2017visual}
Paul~Hongsuck Seo, Andreas Lehrmann, Bohyung Han, and Leonid Sigal. 2017.
\newblock Visual reference resolution using attention memory for visual dialog.
\newblock \emph{Advances in neural information processing systems}, 30.

\bibitem[{Sharma et~al.(2018)Sharma, Ding, Goodman, and
  Soricut}]{sharma2018conceptual}
Piyush Sharma, Nan Ding, Sebastian Goodman, and Radu Soricut. 2018.
\newblock Conceptual captions: A cleaned, hypernymed, image alt-text dataset
  for automatic image captioning.
\newblock In \emph{Proceedings of the 56th Annual Meeting of the Association
  for Computational Linguistics (Volume 1: Long Papers)}, pages 2556--2565.

\bibitem[{Shuster et~al.(2020)Shuster, Humeau, Bordes, and
  Weston}]{shuster-etal-2020-image}
Kurt Shuster, Samuel Humeau, Antoine Bordes, and Jason Weston. 2020.
\newblock \href {https://doi.org/10.18653/v1/2020.acl-main.219} {Image-chat:
  Engaging grounded conversations}.
\newblock In \emph{Proceedings of the 58th Annual Meeting of the Association
  for Computational Linguistics}, pages 2414--2429, Online. Association for
  Computational Linguistics.

\bibitem[{Singh et~al.(2022)Singh, Hu, Goswami, Couairon, Galuba, Rohrbach, and
  Kiela}]{singh2022flava}
Amanpreet Singh, Ronghang Hu, Vedanuj Goswami, Guillaume Couairon, Wojciech
  Galuba, Marcus Rohrbach, and Douwe Kiela. 2022.
\newblock Flava: A foundational language and vision alignment model.
\newblock In \emph{Proceedings of the IEEE/CVF Conference on Computer Vision
  and Pattern Recognition}, pages 15638--15650.

\bibitem[{Stewart et~al.(2016)Stewart, Andriluka, and
  Ng}]{russell2016hungarianmatcher}
Russell Stewart, Mykhaylo Andriluka, and Andrew~Y. Ng. 2016.
\newblock \href {https://doi.org/10.1109/CVPR.2016.255} {End-to-end people
  detection in crowded scenes}.
\newblock In \emph{2016 IEEE Conference on Computer Vision and Pattern
  Recognition (CVPR)}, pages 2325--2333.

\bibitem[{Sun et~al.(2022)Sun, Wang, Xu, Zheng, Yang, Hu, Xu, Zhang, Geng, and
  Jiang}]{sun-etal-2022-multimodal}
Qingfeng Sun, Yujing Wang, Can Xu, Kai Zheng, Yaming Yang, Huang Hu, Fei Xu,
  Jessica Zhang, Xiubo Geng, and Daxin Jiang. 2022.
\newblock \href {https://doi.org/10.18653/v1/2022.acl-long.204} {Multimodal
  dialogue response generation}.
\newblock In \emph{Proceedings of the 60th Annual Meeting of the Association
  for Computational Linguistics (Volume 1: Long Papers)}, pages 2854--2866,
  Dublin, Ireland. Association for Computational Linguistics.

\bibitem[{Sundar and Heck(2022)}]{sundar-heck-2022-multimodal}
Anirudh Sundar and Larry Heck. 2022.
\newblock \href {https://doi.org/10.18653/v1/2022.nlp4convai-1.12} {Multimodal
  conversational {AI}: A survey of datasets and approaches}.
\newblock In \emph{Proceedings of the 4th Workshop on NLP for Conversational
  AI}, pages 131--147, Dublin, Ireland. Association for Computational
  Linguistics.

\bibitem[{Utescher and Zarrie{\ss}(2021)}]{utescher2021did}
Ronja Utescher and Sina Zarrie{\ss}. 2021.
\newblock What did this castle look like before? exploring referential
  relations in naturally occurring multimodal texts.
\newblock In \emph{Proceedings of the Third Workshop on Beyond Vision and
  LANguage: inTEgrating Real-world kNowledge (LANTERN)}, pages 53--60.

\bibitem[{Vaswani et~al.(2017)Vaswani, Shazeer, Parmar, Uszkoreit, Jones,
  Gomez, Kaiser, and Polosukhin}]{vaswani2017attention}
Ashish Vaswani, Noam Shazeer, Niki Parmar, Jakob Uszkoreit, Llion Jones,
  Aidan~N Gomez, {\L}ukasz Kaiser, and Illia Polosukhin. 2017.
\newblock Attention is all you need.
\newblock \emph{Advances in neural information processing systems}, 30.

\bibitem[{Wang et~al.(2022)Wang, Lu, Li, Tao, Guo, Gong, and
  Liu}]{wang2022cris}
Zhaoqing Wang, Yu~Lu, Qiang Li, Xunqiang Tao, Yandong Guo, Mingming Gong, and
  Tongliang Liu. 2022.
\newblock Cris: Clip-driven referring image segmentation.
\newblock In \emph{Proceedings of the IEEE/CVF Conference on Computer Vision
  and Pattern Recognition}, pages 11686--11695.

\bibitem[{Yu et~al.(2016)Yu, Poirson, Yang, Berg, and Berg}]{yu2016modeling}
Licheng Yu, Patrick Poirson, Shan Yang, Alexander~C Berg, and Tamara~L Berg.
  2016.
\newblock Modeling context in referring expressions.
\newblock In \emph{European Conference on Computer Vision}, pages 69--85.
  Springer.

\end{thebibliography}
\bibliographystyle{acl_natbib}

\clearpage

\appendix

\section{MDETR Architecture}

\begin{figure}[h]
    \centering
    \resizebox{\linewidth}{!}{
        \includegraphics{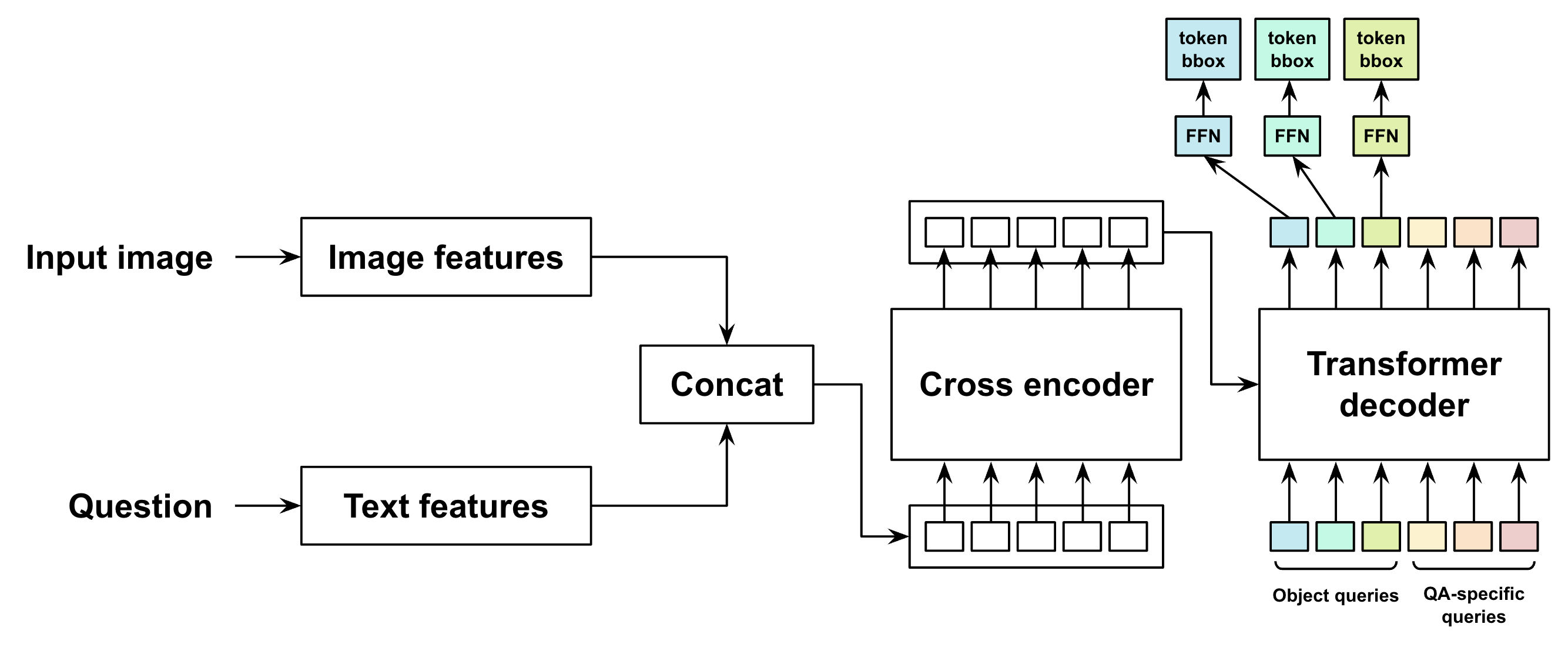}
    }
    \caption{MDETR architecture.}
    \label{fig:mdetr-architecture}
\end{figure}

We provide Figure~\ref{fig:mdetr-architecture} for illustrative comparison with our proposed SitCoM-DETR in \S\ref{sec:contextualized-obj-det}.

\label{sec:appendix}


\end{document}